\documentclass{article} 
\usepackage{iclr2027_conference,times}

\usepackage{amsmath,amsfonts,bm}

\def\eqref#1{equation~\ref{#1}}

\def\1{\bm{1}}

\DeclareMathAlphabet{\mathsfit}{\encodingdefault}{\sfdefault}{m}{sl}
\SetMathAlphabet{\mathsfit}{bold}{\encodingdefault}{\sfdefault}{bx}{n}

\usepackage{hyperref}
\usepackage{url}

\definecolor{boxred}{HTML}{990000}
\definecolor{boxgreen}{HTML}{336600}
\newcommand{\textred}[1]{\textcolor{boxred}{#1}}
\newcommand{\textgreen}[1]{\textcolor{boxgreen}{#1}}

\usepackage{amsmath}
\usepackage{fontawesome5}
\usepackage{pifont}
\newcommand{\xmark}{\ding{55}}
\usepackage{enumitem}
\setlist[itemize]{leftmargin=*}
\usepackage{graphicx}
\usepackage{booktabs}
\usepackage{multirow}

\title{REALM: A Coarse-to-Fine Generative \\Framework for Embodied Reactive Listening}

\author{Peizhen Li\thanks{ Work completed at Macquarie University} ~\& Longbing Cao  \\
Macquarie University, Australia \\
\texttt{peizhen.li@adelaide.edu.au} \\
\texttt{longbing.cao@mq.edu.au} \\
\And
Yang Zhang \\
University of North Texas, USA \\
\texttt{yang.zhang@unt.edu} \\
}

\iclrfinalcopy 
\begin{document}

\maketitle
\addtocontents{toc}{\protect\setcounter{tocdepth}{0}}

\begin{abstract}
Generating responsive listener facial motion is an important task
for embodied conversational AI. Two modeling challenges are central:
accounting for the timing of speaker cues while maintaining continuity
with the listener's ongoing motion, and capturing locally variable
facial events alongside the overall motion trajectory. Listener
responses may follow preceding cues with a temporal lag, while brief
expressions and blinks introduce variation that is difficult to
predict deterministically. These challenges motivate a framework that
combines history-aware temporal alignment with stochastic expression
refinement.
We propose
\textbf{REALM} (\textbf{R}eactive \textbf{E}mbodied
\textbf{A}udio-driven \textbf{L}istening \textbf{M}odel), a
coarse-to-fine framework for audio-driven reactive listening.
A Reactive Gated Speaker--Listener Fusion module combines listener
motion history with speaker audio through a delay-centered attention
prior and adaptive gating. A coarse decoder predicts a base motion
trajectory, which is augmented by audio-conditioned stochastic
residuals in the expression subspace while retaining the coarse pose
parameters. Evaluations on ViCo and L2L show improvements over the
evaluated baselines across multiple motion-quality metrics.
Additional analyses examine delay sensitivity, gate behavior, and
blink dynamics. Finally, deployment on an Ameca humanoid robot and
a perceptual user study demonstrate the applicability of the
generated behavior to physical embodiment.


\faGithub~\textbf{Code:} \href{https://github.com/lipzh5/REALM}{github.com/lipzh5/REALM} \quad \faYoutube~\textbf{Demo:} \href{https://youtu.be/Tf5mpd5S8VQ}{youtu.be/Tf5mpd5S8VQ}
\end{abstract}

\section{Introduction}

Human conversation is a fundamentally dyadic process, driven not only by active speech but also by non-verbal reactive listening~\cite{yngve1970getting, gratch2007creating}. Synthesizing these behaviors—such as head nods, eye contact, and shifts in expression—is vital for signaling comprehension and empathy~\cite{lakin2003chameleon,li2025ugotme}. As a frontier in digital avatar generation, creating realistic listening heads is essential for advancing human-robot interaction, telepresence, and embodied conversational agents~\cite{li2026vividface,urakami2023nonverbal}.

\begin{figure}[h]
\centering\includegraphics[width=1.0\linewidth]{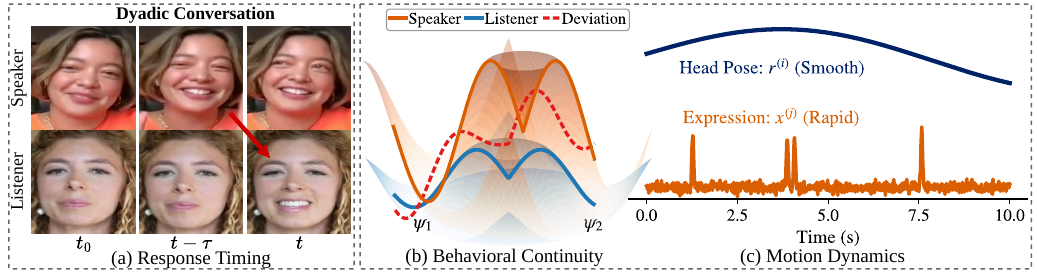}
\caption{\textbf{Motivation for responsive listener motion generation.}
\textbf{(a) Response timing:} Listener responses can follow preceding
speaker cues with a temporal lag.
\textbf{(b) Behavioral continuity:} Speaker activity and listener
motion need not vary proportionally, motivating adaptive use of
speaker evidence and listener history.
\textbf{(c) Motion dynamics:} Overall motion trajectories can coexist
with brief, locally variable facial events, motivating expression
refinement in addition to coarse prediction.}
\label{fig:motivation_issue}
\end{figure}

Listener motion combines periods of limited movement with
intermittent responses to conversational cues. Listener feedback
can also shape the speaker's ongoing
narration~\cite{bavelas2000listeners}. These interactions pose two
modeling challenges. \textbf{First, generation must account for both
the timing of speaker cues and the listener's ongoing motion.}
A response need not coincide with its associated cue, and speaker
activity does not always require a corresponding change in listener
motion (Fig.~\ref{fig:motivation_issue}(a,b)). More broadly, research
on conversational turn-taking highlights the importance of temporal
coordination between interlocutors~\cite{levinson2016turn}.
Recent listener motion provides context for how a response can
develop, while preceding speaker cues provide evidence for its
timing and form. Existing methods explore speaker-conditioned
generation, autoregressive prediction, and dyadic representation
learning~\cite{zhou2022responsive,ng2022learning,liu2024listenformer,tran2024dim}.
Building on these directions, we investigate a delay-aware alignment
prior combined with adaptive weighting of listener history and
speaker evidence.

\textbf{Second, the overall motion trajectory coexists with locally
variable facial dynamics}, including brief expressions and blinks
(Fig.~\ref{fig:motivation_issue}(c)). A conversational context may
admit multiple plausible listener responses, making precise local
motion difficult to predict from audio alone. Deterministic
reconstruction objectives can attenuate this variation by favoring
central tendencies across possible responses. Prior work addresses
this uncertainty through discrete motion representations and
generative modeling~\cite{ng2022learning,tran2024dim,wang2025diffusion}.
We investigate a complementary coarse-to-fine design that preserves
a base trajectory while modeling additional expression variation.
Restricting refinement to expressions allows local stochastic
variation without directly perturbing the coarse pose parameters~\cite{zhou2019talking}.

We propose \textbf{REALM}, a framework for audio-driven reactive
listening that integrates these two design principles. Its Reactive
Gated Speaker--Listener Fusion module uses a shifted ALiBi
bias~\cite{presstrain} to favor speaker representations near a nominal
response lag. This soft prior accommodates content-dependent
alignment, while a learned gate adaptively weights the aligned
speaker context and listener history. The attention bias and gate
thus address complementary aspects of conditioning: temporal
alignment and the relative contribution of each information source.
Together, they provide an inductive bias for balancing continuity
with responsiveness in coarse motion prediction.

The fused context drives a coarse decoder that predicts expression
and pose trajectories. A subsequent refinement module generates
audio-conditioned stochastic expression residuals, leaving the given
coarse pose unchanged. The coarse trajectory provides the base
prediction, while the refinement pathway models residual expression
variation conditioned on that prediction and the speaker audio.
This division allows the two stages to emphasize trajectory
reconstruction and local dynamic variation, respectively.

We evaluate REALM on ViCo and L2L through quantitative comparisons,
component ablations, and analyses of delay sensitivity, gate behavior and blink dynamics. Across both datasets, REALM attains the lowest expression and pose
$L_1$ errors among the evaluated methods, reduces expression
$\text{FID}_{\Delta\text{fm}}$ from 8.71 to 3.91 on ViCo
$\mathcal{D}_{test}$, and is the most preferred method in the robot
user study, using 1.50M parameters. To examine its applicability beyond digital
motion representations, we retarget the generated sequences to an
Ameca humanoid robot using a robot-specific mapping and relative
motion calibration. A perceptual user study complements the
coefficient-space evaluation by assessing the resulting physical
behavior.

Our contributions are:
\begin{itemize}[nosep]
    \item \textbf{Delay-aware reactive fusion.} We combine a
    delay-centered attention prior with adaptive history--audio
    weighting for listener motion generation.
    \item \textbf{Expression-specific stochastic refinement.}
    We introduce a coarse-to-fine architecture that augments a base
    motion trajectory with audio-conditioned expression residuals
    while preserving the coarse pose parameters during refinement.
    \item \textbf{Empirical and embodied evaluation.} We evaluate
    the framework on two conversational benchmarks, characterize
    its generated dynamics, and demonstrate physical deployment
    on Ameca with perceptual evaluation.
\end{itemize}
\section{Related Work}

\textbf{Listening Head Generation.}
Prior work explores different ways of conditioning listener motion
on speaker cues and conversational
context~\cite{zhou2022responsive,liu2024listenformer,tran2024dim,zhu2025infp}.
These approaches include Transformer-based generation and dyadic
representation learning~\cite{huang2022perceptual,liu2024listenformer,tran2024dim},
as well as diffusion-based motion modeling~\cite{wang2025diffusion}.
Several methods explicitly incorporate listener history to support
continuity across generated
motions~\cite{guo2025arig,liu2024customlistener,ng2022learning,ng2023can}.
For example, L2L~\cite{ng2022learning} and
ARIG~\cite{guo2025arig} use autoregressive prediction, while
CustomListener~\cite{liu2024customlistener} introduces a past-guided
generation module. These studies establish the value of modeling
conversational context and motion history. Building on these
directions, REALM combines a delay-centered attention prior with
adaptive weighting of speaker evidence and listener history.
Its coarse-to-fine architecture further models audio-conditioned
stochastic expression residuals while retaining the coarse pose
parameters, providing a complementary approach to response alignment
and local facial variation.  A detailed comparison of modeling choices is provided in
Appendix~\ref{app:comparison_protocol}.

\textbf{Embodied Avatars and Human-Robot Interaction.}
Synthesizing responsive listener motion carries profound implications for affective human-robot interaction~\cite{li2025x2c,li2025ugotme, cao2025humanoid}, as facial dynamics serve as a primary non-verbal communication channel~\cite{mehrabian2017communication,safavi2025facial}. However, most generative avatars remain confined to virtual environments. Physically grounding these generated motions onto humanoid hardware is highly challenging due to the severe domain gap between latent spaces (e.g., 3DMM coefficients) and a robot's strict physical action space~\cite{li2024driving}. Unlike digital avatars, real robotic faces are constrained by mechanical actuation limits, servo inertia, and the nonlinear dynamics of elastic skin~\cite{hu2026learning,lehmann2016head}. We use a robot-specific retargeting pipeline to map generated facial
coefficients to actuator controls, followed by smoothing and
calibration relative to a mechanical neutral configuration.
Deployment on Ameca provides a physical evaluation of the generated
behavior alongside the coefficient-space benchmarks~\cite{chen2021smile,li2026vividface}.

\section{Method}

\subsection{History-Conditioned Generation with a Delay Prior}
\label{sec:method_overview}

\textbf{Problem Formulation.} Let $\mathbf{M}=\{\mathbf{m}_t\}_{t=1}^{T}$ denote the listener motion sequence,
where each frame $\mathbf{m}_t=[\mathbf{x}_t;\mathbf{r}_t]$ contains a
non-rigid expression component $\mathbf{x}_t\in\mathbb{R}^{d_x}$ and a rigid
head-pose component $\mathbf{r}_t\in\mathbb{R}^{d_r}$.
For a target frame $t$, the model conditions on a speaker-audio window $\mathbf{A}_t$ and the
listener's preceding motion history $\mathbf{H}_t$ over $W$ frames, and models
$p_\theta(\mathbf{m}_t\mid\mathbf{A}_t,\mathbf{H}_t)$.

Listener behavior often combines periods of limited motion with intermittent responses to conversational cues. We therefore model two complementary sources of information: recent listener motion provides context for temporal continuity, while speaker audio provides evidence for responsive changes. Because responses need not coincide with the corresponding acoustic cues, we introduce a delay-aware alignment prior. In addition, local facial dynamics, such as blinks, can be less predictable than the overall motion trajectory, motivating a separate stochastic refinement stage.

\begin{figure}[ht]
    \centering
    \includegraphics[width=\linewidth]{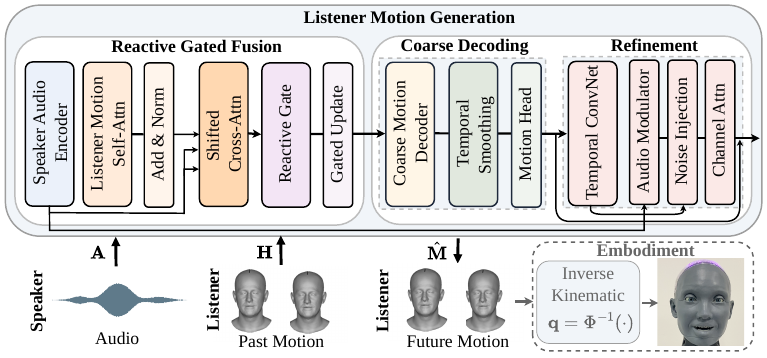}
    \caption{\textbf{Overview of REALM.}
    Speaker audio and listener motion history are integrated through
    delay-aware reactive fusion. The resulting context determines a coarse
    listener trajectory, which is subsequently augmented by
    audio-conditioned stochastic refinement restricted to the expression
    subspace. The generated motion is finally retargeted to the physical robot.}
    \label{fig:system_overview}
\end{figure}

Based on these observations, REALM consists of two corresponding principles.
First, \textbf{Reactive Gated Fusion} introduces a delay-aware prior over
speaker--listener alignment and adaptively balances speaker evidence against
listener history. Second, \textbf{Coarse-to-Fine Stochastic Refinement} first
estimates a coarse motion trajectory and subsequently models residual variation
only in the non-rigid expression subspace. Figure~\ref{fig:system_overview}
illustrates the resulting framework.

\subsection{Reactive Gated Speaker--Listener Fusion}
\label{sec:gated-fusion}

\textbf{Delay-Aware Speaker--Listener Alignment.}
A listener response at time $t$ is not necessarily associated most strongly
with the speaker signal at the same instant. Instead of asking the model to
discover this temporal relationship entirely from data~\cite{hou2019cross}, we introduce a prior
centered on a nominal reaction delay $\tau$.

Let $\mathbf{h}_t$ denote the encoded listener state and $\mathbf{a}_j$ the
speaker representation at time $j$. For attention head $h$, we define the
alignment weights as
\begin{equation}
\pi^{(h)}_{t,j}
\propto
\exp\!\left(
    \frac{
        {\mathbf{q}^{(h)}_t}^{\!\top}\mathbf{k}^{(h)}_j
    }{\sqrt{d_h}}
    -
    m_h |(t-j)-\tau|
\right),
\qquad j\leq t ,
\label{eq:delay_attention}
\end{equation}
where $m_h>0$ controls the strength of the temporal bias, and the weights are normalized over the available speaker indices satisfying $j\leq t$.
The mask excludes future-indexed speaker representations from this attention operation, while the distance term favors a lag near $\tau$. These serve distinct roles: $\tau$ is the center of a soft alignment prior, not a minimum response latency. In particular, indices with $t-\tau<j\leq t$ remain admissible.

Eq.~\ref{eq:delay_attention} can be viewed as combining
\emph{content compatibility} with a \emph{delay-centered temporal prior}:
speaker observations around $t-\tau$ are favored, while the learned
content term remains free to move the effective alignment when supported
by the interaction context. In practice, this formulation is implemented
using our shifted variant of ALiBi~\cite{presstrain}. The resulting
delay-aware speaker context is
$\mathbf{c}^{\tau}_t=\sum_{j\leq t}\pi_{t,j}\mathbf{a}_j$,
with head-specific projections omitted for readability.
This mask defines the temporal restriction at the fusion layer; end-to-end causality additionally depends on the temporal support of the audio representations and subsequent processing. We therefore formulate REALM as window-conditioned generation and do not infer an end-to-end streaming guarantee from this mask alone.

\textbf{Reactive Gating.}
Delay-aware alignment addresses when speaker evidence is preferentially selected, but not how strongly it should influence generation. Listener motion need not change in proportion to ongoing speaker activity. We therefore introduce an adaptive fusion weight~\cite{qiugated} to balance the encoded listener history and the aligned speaker context.

We introduce a learned gate $g_t\in[0,1]$ and represent the
joint context as
\begin{equation}
    \mathbf{z}_t
    =
    \left[
        (1-g_t)\mathbf{h}_t
        \,;\,
        g_t\mathbf{c}^{\tau}_t
    \right].
\label{eq:reactive_fusion}
\end{equation}
Here, $g_t$ controls the relative contribution of the external speaker cue
and the listener's own motion history. A small $g_t$ attenuates the speaker branch and retains more of the history branch; a large $g_t$ increases the relative weight of speaker-conditioned features. The gate is learned jointly with the motion predictor rather than supervised as a reaction detector.

This complementary weighting provides an inductive bias for balancing continuity and responsiveness. While it does not impose a rigid causal constraint on the decoded motion, our empirical examination in Appendix~\ref{app:gate_analysis} confirms that inference-time gate peaks align closely with physical reaction onsets. The coarse listener trajectory is predicted from the resulting temporal context
$\mathbf{Z}=\{\mathbf{z}_t\}$.

\subsection{Coarse-to-Fine Stochastic Refinement}
\label{sec:refine-module}

The fused context in Sec.~\ref{sec:gated-fusion} combines listener history with delay-aware speaker evidence. A separate challenge is how to represent the resulting motion.
A single deterministic predictor must simultaneously account for comparatively
smooth global motion and less predictable facial dynamics. Under reconstruction
objectives, multimodal local variations are easily averaged into a smooth
trajectory; conversely, introducing stochasticity throughout the complete motion
space may unnecessarily perturb stable rigid motion.

We therefore decompose listener generation into a coarse trajectory and a
stochastic residual constrained to the expression subspace. Let
$\tilde{\mathbf{m}}_t=[\tilde{\mathbf{x}}_t;\tilde{\mathbf{r}}_t]$
denote the coarse prediction from the reactive context $\mathbf{Z}$.
We write
\begin{equation}
    \hat{\mathbf m}_t
    =
    \tilde{\mathbf m}_t
    +
    \mathbf P_x \Delta\mathbf x_t,
    \qquad
    \mathbf P_x
    =
    \begin{bmatrix}
        \mathbf I_{d_x}\\
        \mathbf 0
    \end{bmatrix},
\label{eq:coarse_fine}
\end{equation}
where $\mathbf P_x$ embeds an expression residual into the full motion space.
Equivalently,
$\hat{\mathbf m}_t=
[\tilde{\mathbf x}_t+\Delta\mathbf x_t;\tilde{\mathbf r}_t]$.
Thus, stochastic refinement has support only in the non-rigid subspace:
the rigid component remains
$\hat{\mathbf r}_t=\tilde{\mathbf r}_t$ by construction.
This is a separation of output subspaces: the refinement stage leaves the given coarse pose unchanged. It does not impose a strict frequency separation between coarse and residual expression motion.

\textbf{Audio-Conditioned Stochastic Residual.}
The residual should capture variation that is not represented by the coarse
trajectory, while remaining conditioned on the interaction context.
Let $\mathbf z_t^c$ denote the latent representation of the coarse expression.
We define the refinement latent as
\begin{equation}
    \mathbf z_t^r
    =
    \mathbf z_t^c
    +
    \boldsymbol{\beta}_t
    +
    \boldsymbol{\gamma}_t\odot\boldsymbol{\epsilon}_t,
    \qquad
    \boldsymbol{\epsilon}_t\sim
    \mathcal N(\mathbf 0,\mathbf I),
\label{eq:conditional_stochasticity}
\end{equation}
where $(\boldsymbol{\gamma}_t,\boldsymbol{\beta}_t)$ are functions of the
speaker context.
Conditioned on the coarse state and audio, Eq.~\ref{eq:conditional_stochasticity}
induces
\begin{equation}
    \mathbb E[\mathbf z_t^r]
    =
    \mathbf z_t^c+\boldsymbol{\beta}_t,
    \qquad
    \operatorname{Cov}[\mathbf z_t^r]
    =
    \operatorname{diag}(\boldsymbol{\gamma}_t^2).
\label{eq:refinement_moments}
\end{equation}
Hence, $\boldsymbol{\beta}_t$ controls the conditional latent mean shift, whereas $\boldsymbol{\gamma}_t$ controls the latent noise scale. These moments describe the latent distribution before nonlinear decoding; the statistics of the resulting expression residual depend on the learned refinement function.

A refinement function
$\mathcal R_\theta$ maps this stochastic latent to the expression residual,
$\Delta\mathbf x_t=[\mathcal R_\theta(\mathbf Z^r)]_t$, where
$\mathbf Z^r=\{\mathbf z_s^r\}_s$ is the refinement latent sequence. This sequence notation accounts for the temporal processing within the refinement module.
Combining this mapping with Eq.~\ref{eq:coarse_fine} yields
\begin{equation}
    \hat{\mathbf m}_t
    =
    \tilde{\mathbf m}_t
    +
    \mathbf P_x
    \big[\mathcal R_\theta(\mathbf Z^r)\big]_t,
    \qquad
    \mathbf Z^r=\mathbf Z^c+\boldsymbol{\beta}
    +\boldsymbol{\gamma}\odot\boldsymbol{\epsilon}.
\label{eq:realm_refinement}
\end{equation}
Eq.~\ref{eq:realm_refinement} makes the coarse-to-fine decomposition explicit: the coarse pathway supplies the base trajectory, while the refinement pathway models audio-conditioned expression residuals. This parameterization allows multiple expression realizations for the same conditioning input. Whether the learned model uses this stochastic capacity to recover plausible local dynamics is evaluated empirically through the motion and blink analyses.

\textbf{Learning the Decomposition.}
We optimize the coarse and refinement pathways sequentially: the coarse stage prioritizes stable macro-motion, while the
refinement stage focuses on recovering local dynamic variation.
The complete training objectives, including reconstruction, temporal
regularization, and adversarial losses, are provided in
Appendix~\ref{app:extended_implementation}.

\subsection{Physical Grounding and Robotic Embodiment}
\label{sec:robotic-embodiment}

The generated motion $\hat{\mathbf m}_t$ lies in a human facial representation,
whereas physical execution requires commands in the robot action space
$\mathcal Q\subset\mathbb R^{d_q}$. We therefore ground the generated motion
through two deterministic steps: motion retargeting and relative motion calibration.

\textbf{Inverse Kinematic Mapping.}
We denote the mapping from generated facial motion to robot-space controls by
$\Phi^{-1}$. At each time step, we compute $\mathbf q_t = \Phi^{-1} ( \mathbf W\hat{\mathbf m}_t+\mathbf b )$, where $\mathbf W\in\mathbb R^{d_q\times(d_x+d_r)}$ specifies the semantic correspondence between facial components and robot actuators, and $\mathbf b$ contains fixed offsets. The operator $\Phi^{-1}$ additionally applies the hardware-specific constraints required for execution, including actuator-range clipping and the resolution of overlapping control semantics. The complete mapping is specified in Appendix~\ref{app:inverse_mapping}.

\textbf{Relative Motion Calibration.}
Even after mapping into the same action space, absolute human-derived controls cannot be transferred directly because the human representation and robot have different neutral configurations. We therefore preserve relative motion rather than absolute position~\cite{mori2012uncanny, guo2024liveportrait,savitzky1964smoothing}.

Let $\mathcal S(\cdot)$ denote temporal smoothing and define $\tilde{\mathbf q}_t=\mathcal S(\mathbf q_t)$. Given a reference configuration $\tilde{\mathbf q}_{\rm ref}$ extracted from the mapped sequence, its relative motion is $\Delta\mathbf q_t = \tilde{\mathbf q}_t - \tilde{\mathbf q}_{\rm ref}$. Let $\mathbf q_0\in\mathcal Q$ denote the predefined mechanical neutral state of the robot. The calibrated control target is computed as $\hat{\mathbf q}_t = \mathbf q_0 + \Delta\mathbf q_t = \mathbf q_0 + ( \tilde{\mathbf q}_t-\tilde{\mathbf q}_{\rm ref} )$. 

Thus, the mapping determines \emph{which} robot controls correspond to the generated facial motion, while calibration transfers only their displacement from a neutral state. This expresses the smoothed motion relative to the robot's mechanical neutral configuration.

\section{Experiments}
\label{sec:experiments}

\textbf{Experimental Setup and Baselines.} 
We train and evaluate REALM on two conversational benchmarks: \textbf{ViCo}~\cite{zhou2022responsive} (evaluated on in-domain $\mathcal{D}_{test}$ and out-of-domain $\mathcal{D}_{ood}$ splits) and \textbf{L2L}~\cite{ng2022learning}. We evaluate predictions along four axes: (1) \textit{Point-wise Accuracy} ($L_1$ error for expression and pose); (2) \textit{Distributional Realism} (Fréchet Distance, $\text{FID}_{\text{fm}}$, and inter-frame $\text{FID}_{\Delta \text{fm}}$~\cite{yu2023talking}); (3) \textit{Speaker--Listener Correlation} (residual Pearson Correlation Coefficient, rPCC~\cite{tran2024dim}); and (4) \textit{Motion Variability} (temporal variance, var, targeting Ground Truth values). We benchmark against five representative methods: RLHG~\cite{zhou2022responsive}, DSPN~\cite{yu2023responsive}, L2L~\cite{ng2022learning}, ListenFormer~\cite{liu2024listenformer}, and UniLS~\cite{chu2026unils}. Implementation parameters, loss definitions, and evaluation details are provided in Appendix~\ref{app:extended_implementation}.

\subsection{Quantitative Results}
Tables~\ref{tab:quantitative_results_vico} and~\ref{tab:l2l_results} report performance across ViCo and L2L.

\begin{table*}[ht]
\centering
\caption{Quantitative evaluation of REALM against state-of-the-art baselines on the ViCo dataset. We report point-wise accuracy ($L_1$), distributional realism
(FD, $\text{FID}_{\text{fm}}$, and $\text{FID}_{\Delta\text{fm}}$),
speaker--listener correlation (rPCC), and motion variability (var). $L_1$, $\text{FID}_{\Delta \text{fm}}$, and pose FD are scaled by 100. $\downarrow$ denotes lower is better. For var, values closer to the Ground Truth (GT) are better. \textbf{Bold} indicates the best result.}
\label{tab:quantitative_results_vico}
\resizebox{\textwidth}{!}{
\begin{tabular}{ll cc cc cc cc cc cc}
\toprule
\multirow{2}{*}{\textbf{Method}} & \multirow{2}{*}{\textbf{Test set}} & \multicolumn{2}{c}{{$\mathbf{L_1}$} $\downarrow$} & \multicolumn{2}{c}{\textbf{FD} $\downarrow$} & \multicolumn{2}{c}{\textbf{FID}{$_{\text{fm}}$} $\downarrow$} & \multicolumn{2}{c}{\textbf{FID}$_{\Delta \text{fm}}$ $\downarrow$} & \multicolumn{2}{c}{\textbf{rPCC} $\downarrow$} & \multicolumn{2}{c}{\textbf{var} $\rightarrow$ GT} \\
\cmidrule(lr){3-4} \cmidrule(lr){5-6} \cmidrule(lr){7-8} \cmidrule(lr){9-10} \cmidrule(lr){11-12} \cmidrule(lr){13-14} 
& & \textbf{exp} & \textbf{pose} & \textbf{exp} & \textbf{pose} & \textbf{exp} & \textbf{pose} & \textbf{exp} & \textbf{pose} & \textbf{exp} & \textbf{pose} & \textbf{exp} & \textbf{pose} \\
\midrule
\multirow{2}{*}{RLHG} & $\mathcal{D}_{test}$ & 14.80 & 8.27 & 0.69 & \textbf{0.72} & 3.19 & 0.084 & 9.88 & 0.89 & 0.039 & 0.024 & 0.113 & 0.150 \\
 & $\mathcal{D}_{ood}$ & 19.75 & 7.37 & 3.27 & 1.69 & 5.86 & 0.074 & 8.16 & 0.84 & 0.065 & 0.092 & 0.115 & 0.195 \\
\midrule
\multirow{2}{*}{DSPN} & $\mathcal{D}_{test}$ & 15.48 & 7.40 & 0.65 & 0.95 & 3.05 & 0.078 & 8.91 & 0.73 & 0.035 & 0.041 & .140 & 0.240 \\
 & $\mathcal{D}_{ood}$ & 24.11 & 8.66 & 2.85 & 1.45 & 5.12 & 0.065 & 6.07 & 0.47 & 0.072 & 0.075 & {0.141} & 0.040 \\
\midrule
\multirow{2}{*}{L2L} & $\mathcal{D}_{test}$ & 20.36 & 10.70 & 0.58 & 1.25 & 3.84 & 0.096 & 12.06 & 0.60 & 0.040 & 0.070 & 0.097 & 0.033 \\ 
 & $\mathcal{D}_{ood}$ & 23.82 & 7.77 & 1.72 & 1.29 & 4.19 & 0.056 & 8.67 & 0.51 & 0.080 & 0.060 & 0.098 & 0.036 \\
\midrule
\multirow{2}{*}{ListenFormer} & $\mathcal{D}_{test}$ & 11.92 & 7.04 & 0.57 & 1.09 & 2.04 & 0.064 & 8.71 & 0.75 & 0.032 & {0.017} & \textbf{0.142} & {0.043} \\
 & $\mathcal{D}_{ood}$ & 17.72 & 5.93 & 2.69 & 0.66 & 4.40 & 0.048 & 6.16 & 0.63 & 0.059 & 0.009 & 0.133 & {0.047} \\
\midrule
\multirow{2}{*}{UniLS} & $\mathcal{D}_{test}$ & 25.50 & 9.45 & 1.89 & 1.06 & 9.01 & 0.092 & 4.48 & 0.13 & 0.284 & 0.132 & 0.130 & 0.032 \\
 & $\mathcal{D}_{ood}$ & 27.15 & 6.82 & 5.71 & {0.57} & 10.25 & 0.036 & 4.92 & {0.18} & 0.062 & 0.037 & 0.114 & 0.026 \\
\midrule
\multirow{2}{*}{\textbf{REALM (Ours)}} & $\mathcal{D}_{test}$ & \textbf{11.58} & \textbf{6.52} & \textbf{0.56} & 1.01 & \textbf{1.93} & \textbf{0.050} & \textbf{3.91} & \textbf{0.09} & \textbf{0.009} & 
\textbf{0.014} & 0.133 & \textbf{0.045} \\
 & $\mathcal{D}_{ood}$ & \textbf{13.76} & \textbf{5.02} & \textbf{1.47} & \textbf{0.52} & \textbf{2.79} & \textbf{0.030} & \textbf{4.67} & \textbf{0.15} & \textbf{0.050} & \textbf{0.005} & \textbf{0.145} & 
 \textbf{0.051} \\
\bottomrule
\end{tabular}
}
\end{table*}

\textbf{Point-wise Accuracy and Distributional Realism.} 
REALM consistently achieves the lowest point-wise $L_1$ errors across both datasets for expression and pose. In terms of static distribution metrics ($\text{FD}$ and $\text{FID}_{\text{fm}}$), REALM achieves competitive expression realism on ViCo $\mathcal{D}_{test}$ (0.56 vs. 0.57 for ListenFormer) and notable gains under out-of-domain evaluation ($\mathcal{D}_{ood}$ expression FD of 1.47 vs. 1.72--5.71 for baselines). For rigid pose distribution on $\mathcal{D}_{test}$, however, RLHG retains a lower pose FD (0.72 vs. 1.01). 
The most pronounced difference appears in the temporal transition metric ($\text{FID}_{\Delta\text{fm}}$): REALM achieves an expression score of 3.91 on $\mathcal{D}_{test}$ and 6.50 on L2L, improving upon ListenFormer (8.71 and 13.41). This suggests that incorporating stochastic refinement primarily aids inter-frame facial dynamics rather than static trajectory tracking. Interestingly, while UniLS exhibits higher spatial tracking error ($L_1$ of 25.50 on $\mathcal{D}_{test}$), it maintains strong dynamic smoothness ($\text{FID}_{\Delta\text{fm}}$ of 4.48), demonstrating the distinct trade-offs between static coordinate fidelity and dynamic continuity.

\textbf{Speaker--Listener Correlation and Motion Variability.} 
REALM demonstrates close alignment with ground-truth interaction dynamics, yielding the lowest rPCC values across both benchmarks. On L2L, ListenFormer shows comparable correlation performance (rPCC of 0.008 / 0.010 vs. 0.007 / 0.008 for REALM). 
Regarding motion variability (var), REALM closely matches ground-truth variance levels on ViCo $\mathcal{D}_{ood}$ and L2L. On ViCo $\mathcal{D}_{test}$, ListenFormer achieves an expression variance (0.142) slightly closer to the reference distribution than REALM (0.133), indicating that strong deterministic architectures can maintain global movement magnitude, even if high-frequency micro-dynamics remain less varied.

\begin{table*}[t]
\centering
\caption{Quantitative evaluation on the L2L dataset. Evaluation metrics, scaling factors, and formatting conventions are identical to Table \ref{tab:quantitative_results_vico}.}
\label{tab:l2l_results}
\resizebox{\textwidth}{!}{
\begin{tabular}{l cc cc cc cc cc cc}
\toprule
\multirow{2}{*}{\textbf{Method}} & \multicolumn{2}{c}{{$\mathbf{L_1}$} $\downarrow$} & \multicolumn{2}{c}{\textbf{FD} $\downarrow$} & \multicolumn{2}{c}{\textbf{FID}$_{\text{fm}}$ $\downarrow$} & \multicolumn{2}{c}{\textbf{FID}$_{\Delta \text{fm}}$ $\downarrow$} & \multicolumn{2}{c}{\textbf{rPCC} $\downarrow$} & \multicolumn{2}{c}{\textbf{var} $\rightarrow$ GT} \\
\cmidrule(lr){2-3} \cmidrule(lr){4-5} \cmidrule(lr){6-7} \cmidrule(lr){8-9} \cmidrule(lr){10-11} \cmidrule(lr){12-13}
& \textbf{exp} & \textbf{pose} & \textbf{exp} & \textbf{pose} & \textbf{exp} & \textbf{pose} & \textbf{exp} & \textbf{pose} & \textbf{exp} & \textbf{pose} & \textbf{exp} & \textbf{pose} \\
\midrule
RLHG   & 17.49    & 4.92   & 63.88 & 0.87 & 3.19     & 0.037       & 11.38      & 0.052     & 0.102     & 0.063        & 0.153      & 0.006      \\
DSPN           & 23.65    & 5.82    & 72.15 & 1.12 & 3.85     & 0.045       & 12.40      & 0.085     & 0.120     & 0.082       & 0.135      & 0.005      \\
L2L          & 10.45    & 2.88    & 14.63 & 0.11 & 1.81     & 0.022       & 11.32     & 0.079       & 0.026      & 0.034       & 0.124      & 0.011      \\
ListenFormer  & 10.32    & 2.57    & 17.12    & 0.07    & 1.44 & 0.016 & 13.41    & 0.051       & 0.008      & 0.010       & 0.173      & 0.010      \\
UniLS & 26.10 & 4.50 & 95.00 & 0.65 & 7.50 & 0.032 & 11.80 & 0.055 & 0.300 & 0.120 & 0.150 & 0.007 \\
\midrule
\textbf{REALM (Ours)} & \textbf{9.67} & \textbf{2.41} & \textbf{9.85} & \textbf{0.05} & \textbf{1.23} & \textbf{0.013} & \textbf{6.50}  & \textbf{0.048} & \textbf{0.007} & \textbf{0.008} & \textbf{0.181} & \textbf{0.012} \\
\bottomrule
\end{tabular}
}
\end{table*}

\textbf{Model Ablation.}
Evaluating REALM components on ViCo $\mathcal{D}_{test}$ (Table~\ref{tab:ablation_study_expanded}) highlights their specific contributions. Shifted Attention (SA) aligns listener reaction timing with speaker cues, reducing pose rPCC from 0.026 to 0.018. However, omitting Gated Fusion (GF) allows the model to over-react to acoustic energy, leading to structural drift (expression FD rises to 0.68 with SA alone). Combining GF with SA stabilizes the base trajectory, yielding the lowest pose error ($L_1$ of 6.52 and FD of 1.01). Meanwhile, the Refinement Module (RM) injects audio-conditioned stochastic variation to cure deterministic over-smoothing, improving expression FD to 0.62 in isolation and 0.56 in the full model. Because RM is constrained to the non-rigid expression subspace ($\hat{\mathbf{r}}_t = \tilde{\mathbf{r}}_t$), pose kinematics are unaffected; instead, RM drives dynamic facial fidelity, sharply reducing expression $\text{FID}_{\Delta\text{fm}}$ from $\sim 12.50$ down to 4.11 on its own and 3.91 when integrated. A blink analysis in Appendix~\ref{app:blink_analysis} verifies that these recovered micro-dynamics correspond to structured anatomical motions rather than high-frequency noise.

\begin{table*}[ht]
\centering
\caption{Ablation study of REALM components on the ViCo ($\mathcal{D}_{test}$) split evaluating Gated Fusion (GF), the Refinement Module (RM), and Shifted Attention (SA). Because RM refines only non-rigid expressions ($\hat{\mathbf{r}}_t = \tilde{\mathbf{r}}_t$), pose metrics are determined entirely by the coarse stage (GF and SA). Metrics follow Table~\ref{tab:quantitative_results_vico}.}
\label{tab:ablation_study_expanded}
\resizebox{\textwidth}{!}{%
\begin{tabular}{ccc cc cc cc cc cc cc}
\toprule
\multicolumn{3}{c}{\textbf{Modules}} & \multicolumn{2}{c}{{$\mathbf{L_1}$} $\downarrow$} & \multicolumn{2}{c}{\textbf{FD} $\downarrow$} & \multicolumn{2}{c}{\textbf{FID}$_{\text{fm}}$ $\downarrow$} & \multicolumn{2}{c}{\textbf{FID}$_{\Delta \text{fm}}$ $\downarrow$} & \multicolumn{2}{c}{\textbf{rPCC} $\downarrow$} & \multicolumn{2}{c}{\textbf{var} $\rightarrow$ GT} \\
\cmidrule(lr){1-3} \cmidrule(lr){4-5} \cmidrule(lr){6-7} \cmidrule(lr){8-9} \cmidrule(lr){10-11} \cmidrule(lr){12-13} \cmidrule(lr){14-15}
GF & RM & SA & \textbf{exp} & \textbf{pose} & \textbf{exp} & \textbf{pose} & \textbf{exp} & \textbf{pose} & \textbf{exp} & \textbf{pose} & \textbf{exp} & \textbf{pose} & \textbf{exp} & \textbf{pose} \\
\midrule
\xmark & \xmark & \xmark & 12.20 & 6.69 & 0.64 & 1.19 & 2.27 & 0.060 & 12.51 & 0.10 & 0.014 & 0.026 & 0.132 & 0.042 \\
\checkmark & \xmark & \xmark & 12.04 & 6.78 & 0.63 & 1.03 & 2.28 & 0.060 & 12.50 & 0.10 & 0.019 & 0.016 & 0.132 & 0.043 \\
\xmark & \checkmark & \xmark & 12.25 & 6.69 & 0.62 & 1.19 & 2.18 & 0.060 & 4.11 & 0.10 & 0.012 & 0.026 & 0.131 & 0.042 \\
\xmark & \xmark & \checkmark & 11.98 & \textbf{6.52} & 0.68 & 1.32 & 2.19 & 0.060 & 12.53 & 0.10 & 0.012 & 0.018 & 0.131 & 0.041 \\
\midrule
\xmark & \checkmark & \checkmark & 12.31 & \textbf{6.52} & 0.60 & 1.32 & 2.09 & 0.060 & 4.10 & 0.10 & 0.011 & 0.018 & 0.130 & 0.041 \\
\checkmark & \checkmark & \xmark & 12.09 & 6.78 & 0.67 & 1.03 & 2.16 & 0.060 & 4.13 & 0.10 & 0.023 & 0.016 & 0.132 & 0.043 \\
\checkmark & \xmark & \checkmark & 11.85 & \textbf{6.52} & 0.59 & \textbf{1.01} & 2.22 & \textbf{0.050} & 12.49 & \textbf{0.09} & 0.015 & \textbf{0.014} & 0.131 & \textbf{0.045} \\
\checkmark & \checkmark & \checkmark & \textbf{11.58} & \textbf{6.52} & \textbf{0.56} & \textbf{1.01} & \textbf{1.93} & \textbf{0.050} & \textbf{3.91} & \textbf{0.09} & \textbf{0.009} & \textbf{0.014} & \textbf{0.133} & \textbf{0.045} \\
\bottomrule
\end{tabular}%
}
\end{table*}

\textbf{Parameter Analysis of the Delay Prior ($\tau$).}
We examine the sensitivity of REALM to the nominal delay $\tau$ in the
shifted attention bias on the ViCo dataset. At 30 FPS
(1 frame $\approx$ 33.3 ms), we evaluate shifts of
$\tau \in \{0,4,8,12\}$ frames
(Table~\ref{tab:reaction_delay}).
Among the tested settings, an 8-frame shift ($\approx$ 267 ms)
achieves the lowest $L_1$, FD, and rPCC errors for both expressions
and head poses. Relative to the unshifted setting, these results
support the usefulness of favoring preceding speaker cues when
generating listener responses. Increasing the shift to 12 frames
($\approx$ 400 ms) worsens performance relative to the 8-frame
setting, particularly in rPCC, indicating sensitivity to the
location of the temporal prior.
Importantly, $\tau$ specifies a preferred alignment lag rather
than a fixed response latency: the content-dependent attention
weights can favor other admissible time steps. Accordingly,
this analysis supports the selected delay prior under the evaluated
configuration, rather than establishing a universal physiological
reaction time.

\begin{table}[ht]
\centering
\caption{Sensitivity to the nominal delay $\tau$ in the attention
prior on ViCo at 30 FPS. An 8-frame shift performs best among
the tested settings. Time values indicate the center of the
temporal bias, not a prescribed response latency.}
\label{tab:reaction_delay}
\vspace{5pt}
\small
\begin{tabular}{lcccc}
\toprule
\textbf{Prior Center}
& \textbf{$\tau$ (frames)}
& \textbf{$L_1 \downarrow$ (exp / pose)}
& \textbf{FD $\downarrow$ (exp / pose)}
& \textbf{rPCC $\downarrow$ (exp / pose)} \\
\midrule
0 ms (No Shift)
& 0 & 11.88 / 6.96 & 0.61 / 1.43 & 0.019 / 0.029 \\
$\approx$ 133 ms
& 4 & 11.71 / 6.85 & 0.58 / 1.31 & 0.018 / 0.029 \\
\textbf{$\approx$ 267 ms (Ours)}
& \textbf{8}
& \textbf{11.58 / 6.52}
& \textbf{0.56 / 1.01}
& \textbf{0.009 / 0.014} \\
$\approx$ 400 ms
& 12 & 11.76 / 6.82 & 0.57 / 1.36 & 0.020 / 0.045 \\
\bottomrule
\end{tabular}
\end{table}

\subsection{Robotic Embodiment and User Study}

\begin{figure}[ht]
    \centering
    \includegraphics[width=0.9\linewidth]{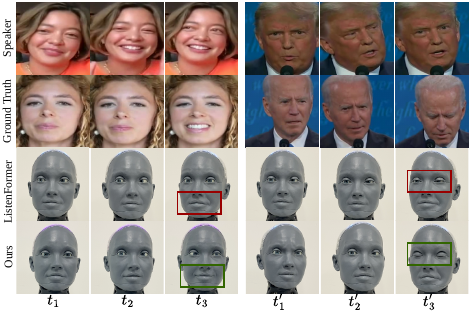}
    \caption{\textbf{Qualitative comparison of physical embodiment on the Ameca robot.} To illustrate the performance observed across the ViCo test set, we visualize two representative video sequences (frames $t_1, t_2, t_3$ and $t_1', t_2', t_3'$) comparing the reactive listener motion generated by our method (REALM) against the human Ground Truth and the ListenFormer~\cite{liu2024listenformer} baseline. Driven by the speaker's context (Top Row), ListenFormer frequently suffers from deterministic over-smoothing, failing to synthesize intended expressions such as a conversational smile or a natural blink (highlighted in \textred{red}). In contrast, REALM effectively anchors the listener to their natural reactive manifold, accurately recovering contextually appropriate expressions and subtle stochastic micro-dynamics (e.g., smiling and blinking, highlighted in \textgreen{green}) that closely match the human Ground Truth.}
    \label{fig:robot_exp}
\end{figure}

\textbf{Qualitative Analysis.} While latent metrics evaluate mathematical fidelity, they often fail to capture physical plausibility. To evaluate the complete embodied pipeline, we mapped the generated motions from the ViCo test set into physical, robot-ready control values using our inverse kinematic mapping. Figure~\ref{fig:robot_exp} presents a visual comparison against the human ground truth and ListenFormer~\cite{liu2024listenformer}. We specifically select ListenFormer for this visual analysis because it represents the most recent and strongest baseline among our evaluated methods. As demonstrated in the visualized sequences, baselines lacking stochastic refinement frequently suffer from deterministic over-smoothing. They struggle to deviate from average mean poses, failing to synthesize intended expressions such as a conversational smile or a rapid, natural blink. In contrast, by coupling our gating mechanism with an audio-conditioned refinement module, REALM effectively anchors the listener within their natural reactive manifold. This allows our model to maintain stable, contextually appropriate resting states while successfully recovering the high-frequency micro-dynamics required to match the perceptual vividness of the human ground truth.

\textbf{User Study.}
We conducted a Mean Opinion Score (MOS) user
study~\cite{cudeiro2019capture} to assess the perceived quality
of the generated behavior on Ameca. Each of $N=25$ evaluators
viewed five conversational sequences from the ViCo test set.
For each sequence, participants evaluated REALM and five
baselines: RLHG, DSPN, L2L, ListenFormer, and UniLS.
Method identities were concealed, and their display order
was randomized. Drawing on established evaluation
frameworks~\cite{bartneck2009measurement}, participants rated
each method on a 1--5 scale across four criteria:
Naturalness, Audio-Visual Synchrony, Contextual Appropriateness,
and Overall Preference. Attention-check questions were embedded 
to ensure response reliability, and pairwise score differences 
between REALM and the strongest baseline were statistically 
significant ($p < 0.01$, Wilcoxon signed-rank test). Further 
procedural details and criterion definitions are provided in
Appendix~\ref{app:user_study}.

\begin{table}[htbp]
    \vspace{2mm}
    \noindent 
    \begin{minipage}[t]{0.45\textwidth}
        \vspace{0pt} 
        \textbf{Analysis of Robot Embodiment:} As detailed in Table~\ref{tab:user_study}, our proposed approach consistently outperforms prior methods during physical deployment on the Ameca platform. Baselines such as ListenFormer frequently exhibit deterministic smoothing or inappropriate mirroring, leading to lower Contextual Appropriateness (App.) and Naturalness (Nat.) during prolonged dyadic exchanges. 
        In contrast, our reactive architecture prevents temporal drift, resulting in superior Audio-Visual Synchrony (Syn.) and achieving the highest Overall Preference (Pref.) among human evaluators.
    \end{minipage}\hfill
    \begin{minipage}[t]{0.52\textwidth}
        \vspace{0pt} 
        \vspace{-\abovecaptionskip} 
        \caption{User study results on Ameca, reported as mean opinion scores
on a 1--5 scale ($N=25$; five sequences per participant; six methods
per sequence). Higher scores indicate better perceived quality.
\textbf{Bold} indicates the highest mean score.}
        \label{tab:user_study}
        \vspace{5pt}
        \small 
        \begin{tabular}{@{}lcccc@{}}
        \toprule
        \textbf{Method} & \textbf{Nat.} $\uparrow$ & \textbf{Syn.} $\uparrow$ & \textbf{App.} $\uparrow$ & \textbf{Pref.} $\uparrow$ \\
        \midrule
        RLHG & 2.2 & 2.4 & 2.4 & 2.2 \\
        DSPN & 2.4 & 2.6 & 2.8 & 2.4 \\
        L2L & 2.8 & 3.0 & 3.2 & 2.8 \\
        ListenFormer & 3.6 & 3.8 & 4.0 & 3.6 \\
        UniLS & 3.2 & 3.6 & 3.8 & 3.4 \\
        \midrule
        \textbf{REALM (Ours)} & \textbf{4.4} & \textbf{4.2} & \textbf{4.2} & \textbf{4.0} \\
        \bottomrule
        \end{tabular}
    \end{minipage}
\end{table}

\section{Conclusion and Future Work}
\label{sec:conclusion}

We presented REALM, a coarse-to-fine framework for audio-driven
listener motion generation. Its fusion module combines listener
history with speaker context through a delay-centered attention
prior and adaptive gating. Its refinement module adds
audio-conditioned stochastic expression residuals while preserving
the coarse head-pose trajectory. Evaluations on ViCo and L2L
demonstrate improvements over the evaluated baselines, with
additional analyses examining delay sensitivity, gate behavior,
and blink dynamics. Deployment on an Ameca robot and a perceptual
user study further demonstrate the applicability of the generated
motion to physical embodiment.

\textbf{Limitations and Future Work.}
REALM currently operates on offline conversational windows;
extending it to streaming human--robot interaction requires
explicit control of temporal information access and end-to-end
latency. Furthermore, the nominal delay provides a shared alignment prior
that may not capture variation across individuals, making the development of 
context-dependent delay priors a promising direction. Finally, our physical
deployment relies on a robot-specific retargeting pipeline; extending
this approach across different embodiments will require
morphology-aware mappings and validation of actuator limits
on each respective platform.

\newpage
\section*{AI Use Statement}
In this work, we used generative AI tools to polish parts of the research paper to improve readability, and to help refine software code by identifying and fixing bugs. We have not used generative AI tools to generate synthetic data sets, develop theoretical models, formulate mathematical claims, design experiments, or interpret results. We have thoroughly reviewed all AI-assisted work. Specifically, all AI-suggested code fixes were manually verified and tested for correctness within our PyTorch pipeline, and all polished text was reviewed by the authors to ensure the original scientific meaning and claims were strictly preserved. The authors take full responsibility for the final content of this work, including text, claims, or artifacts produced with the aid of generative AI.

\section*{Ethics Statement}
This research focuses on generating reactive listener motions for dyadic conversational interactions and emphasizes both physical and digital safety. First, all datasets utilized in this work (ViCo and L2L) are publicly available, sourced from prior peer-reviewed publications, and contain data that is strictly used in accordance with their respective academic licenses. Second, to ensure physical safety during our embodied human-robot interaction experiments, our physical grounding pipeline deliberately avoids unconstrained neural end-to-end control. Instead, we utilize a deterministic Inverse Kinematics (IK) mapping with hard-coded mechanical safety bounds to strictly prevent hardware collisions or unpredictable physical actuation. Finally, the subjective human evaluation (Mean Opinion Score) was conducted following standard ethical guidelines for observational user studies, involving double-blinded, anonymized video reviews without collecting personally identifiable information (PII) from the $N=25$ participants. We do not foresee any immediate negative societal impacts from this specific responsive listening framework, as it is designed to enhance cooperative human-robot communication rather than generate deceptive or synthetic speaker identities.

\section*{Reproducibility Statement}
We are committed to ensuring the full reproducibility of the REALM framework. To allow the community to verify and build upon our results, we have provided our complete source code and training scripts as a zipped archive within the supplementary materials. Because exhaustive technical specifications can distract from the core conceptual narrative, we have meticulously documented all implementation details in the appendix. Specifically, Appendix \ref{app:extended_implementation} contains the exact mathematical formulations for our two-stage curriculum loss, comprehensive architectural hyperparameters for all core modules, dataset processing protocols, and formal definitions for all evaluation metrics used in Section \ref{sec:experiments}. Furthermore, the empirical inference-time gating data ($\mathbf{g}$) and structural blink analyses are fully detailed in Appendices \ref{app:blink_analysis} and \ref{app:gate_analysis}.

\bibliography{iclr2027_conference}
\bibliographystyle{iclr2027_conference}


\newpage
\appendix
\renewcommand{\contentsname}{APPENDIX}
\addtocontents{toc}{\protect\setcounter{tocdepth}{2}}
\tableofcontents
\newpage


\section{Empirical Analysis of the Gating Mechanism}
\label{app:gate_analysis}

Our fusion gate adaptively weights speaker context and listener history. To examine whether its learned activations are associated with listener motion, we recorded gate values during inference on the ViCo test set. We analyze their alignment with velocity-defined motion onsets and compare ground-truth motion velocity between frames with relatively high and low gate values within each clip. As summarized in Table~\ref{tab:gate_analysis}, we evaluate the mechanism across two primary criteria: its strict temporal alignment with true physical reaction onsets, and its statistical correlation with physical motion velocity.

\begin{table}[h]
\centering
\caption{Inference-time quantitative analysis of the gating mechanism ($\mathbf{g}$) across the ViCo test set.}
\vspace{5pt}
\renewcommand{\arraystretch}{1.2}
\small 
\begin{tabular}{@{}llc@{}}
\toprule
\textbf{Analysis Category} & \textbf{Evaluation Metric} & \textbf{Result} \\
\midrule
\multirow{3}{*}{\begin{tabular}{@{}l@{}}\textbf{Temporal Alignment} \\ ($\mathbf{g}$ Peak vs. GT Onset)\end{tabular}} 
 & Total Reactions & 515 \\
 & Mean Offset & -0.50 frames (-16.6 ms) \\
 & Median Offset & 0.00 frames \\
\midrule
\multirow{2}{*}{\begin{tabular}{@{}l@{}}\textbf{Intra-Clip Velocity} \\ (Top 10\% vs. Bottom 10\% $\mathbf{g}$)\end{tabular}} 
 & Mean $\Delta V$ & 0.0051 \\
 & Statistical Significance & $p < 0.001$ ($3.2 \times 10^{-5}$) \\
\bottomrule
\end{tabular}
\label{tab:gate_analysis}
\end{table}

\subsection{Temporal Alignment with True Reaction Onsets}

To explicitly verify our claim regarding temporal alignment, we established ground-truth reaction onsets by calculating the physical velocity of the human listener. A reaction onset was defined as the exact frame where this velocity crossed a sequence-specific threshold (mean + 0.5 std). We then measured the offset between these physical onsets and the nearest peak in the predicted $\mathbf{g}$ sequence. 
Across 515 velocity-defined motion onsets, the nearest gate peak has a median offset of 0 frames and a mean offset of $-0.50$ frames. These descriptive results indicate temporal alignment between gate peaks and the selected motion onsets. Because both quantities are indexed on the listener timeline, this analysis does not estimate the delay between speaker cues and listener responses or isolate the contribution of shifted attention. Moreover, nearest-peak offsets should be interpreted alongside peak density and an appropriate chance-alignment baseline.

\subsection{Intra-Clip Correlation with Motion Velocity}

Because the network learns to use $\mathbf{g}$ as a continuous blending weight, global aggregation across the dataset can wash out sequence-specific baseline shifts (e.g., Sequence A operating naturally between 0.3--0.7, while Sequence B operates between 0.6--0.9). To strictly control for these baseline shifts and isolate the gate's local causal behavior, we conducted an intra-clip analysis. For every clip, we measured the ground-truth physical velocity during frames with the highest predicted reactivity (Top 10\% of $\mathbf{g}$) versus the lowest reactivity (Bottom 10\% of $\mathbf{g}$).
Within clips, ground-truth motion velocity is higher during frames in the top decile of gate values than during frames in the bottom decile, with a reported mean difference of $0.0051$ and $p=3.2\times10^{-5}$. This association is consistent with the gate assigning greater relative weight to speaker-conditioned features during more active listener motion. It does not by itself establish that gate activations cause motion changes or prevent long-horizon drift.

\section{Extended Implementation Details}
\label{app:extended_implementation}

\subsection{Reactive Gated Fusion Parameterization}
\label{app:reactive_fusion}

Section~\ref{sec:gated-fusion} describes REALM in terms of a delay-aware
speaker context $\mathbf c_t^\tau$ and reactivity variable $g_t$.
Here we give their exact network realization.

Given the speaker-audio window $\mathbf A$ and listener-history window
$\mathbf H$, the corresponding encoders produce
$\mathbf E_A,\mathbf E_H\in\mathbb R^{W\times D}$.
For attention head $h$, queries, keys, and values are

\[
\mathbf Q_h=\mathbf E_H\mathbf W_Q^{(h)},\qquad
\mathbf K_h=\mathbf E_A\mathbf W_K^{(h)},\qquad
\mathbf V_h=\mathbf E_A\mathbf W_V^{(h)}.
\]

We implement the delay prior in Eq.~\ref{eq:delay_attention} using a shifted
ALiBi distance matrix
$D_{ij}=|(i-j)-\tau|$ and causal mask
$M_{ij}=0$ for $j\leq i$ and $-\infty$ otherwise. The pre-softmax scores are

\begin{equation}
\mathbf S_h
=
\frac{\mathbf Q_h\mathbf K_h^\top}{\sqrt{d_h}}
-
m_h\mathbf D
+
\mathbf M_{\rm causal}.
\label{eq:appendix_shifted_alibi}
\end{equation}

The corresponding attention weights are
$\mathbf \Pi_h=\operatorname{softmax}(\mathbf S_h)$, and the multi-head
outputs are concatenated and projected to obtain
$\mathbf C_{\rm fused}$.

The reactivity variable used in Eq.~\ref{eq:reactive_fusion} is implemented as

\begin{equation}
\mathbf g
=
\sigma\!\left(
\mathbf W_2
\operatorname{GELU}
(\mathbf W_1\mathbf C_{\rm fused}+\mathbf b_1)
+\mathbf b_2
\right),
\label{eq:appendix_gate}
\end{equation}

where $\mathbf g\in[0,1]^{W\times1}$.

The implementation-level fused sequence context is therefore

\begin{equation}
\mathbf{Z}
=
\left[
(1-\mathbf{g})\odot\mathbf{E}_H
\,;\,
\mathbf{g}\odot\mathbf{C}_{\text{fused}}
\right],
\label{eq:appendix_gated_memory}
\end{equation}

which is the sequence-level matrix realization of Eq.~\ref{eq:reactive_fusion}.



\subsection{Stochastic Refinement Parameterization}
\label{app:refinement_parameterization}

Section~\ref{sec:refine-module} formulates refinement as a
context-conditioned stochastic residual. We now specify the network
parameterization used to instantiate this formulation.

Let $\tilde{\mathbf X}\in\mathbb R^{W\times d_x}$ denote the expression
component of the coarse trajectory. A temporal encoder
$\mathcal F_{\rm temp}$ maps it to
\[
\mathbf Z^c=\mathcal F_{\rm temp}(\tilde{\mathbf X}),
\]
where $\mathcal F_{\rm temp}$ is implemented using a dilated temporal
convolution.

The speaker representation controls the location and scale of the
stochastic residual through
\begin{equation}
[\boldsymbol{\gamma},\boldsymbol{\beta}]
=
\mathcal F_{\rm mod}(\mathbf E_A),
\label{eq:appendix_audio_modulator}
\end{equation}
and the noisy refinement latent is computed as:
\begin{equation}
\mathbf Z^r
=
\mathbf Z^c+
\boldsymbol{\beta}
+
\boldsymbol{\gamma}\odot\boldsymbol{\epsilon},
\label{eq:appendix_noise}
\end{equation}
where $\boldsymbol{\epsilon} \sim \mathcal{N}(\mathbf{0}, \mathbf{I})$, and the time-varying scale and shift parameters $[\boldsymbol{\gamma}, \boldsymbol{\beta}]$ are implicitly broadcast across the latent feature dimension to match the sequence shape of $\mathbf{Z}^c$.

Channel-wise modulation is then applied as
$\mathbf Z^{a}=\mathbf A_c\odot\mathbf Z^r$,
where
\[
\mathbf A_c
=
\sigma\!\left(
\mathbf W_2^c
\operatorname{ReLU}
\left(
\mathbf W_1^c
\operatorname{GAP}(\mathbf Z^r)
\right)
\right).
\]

A temporal self-attention encoder followed by an output projection realizes
the refinement function
$\mathcal R_\theta$ introduced in
Eq.~\ref{eq:realm_refinement}, yielding
$\Delta\mathbf X=\mathcal R_\theta(\mathbf Z^r)$.
The expression residual is finally inserted into the complete motion
through the projection $\mathbf P_x$ defined in
Eq.~\ref{eq:coarse_fine}.

\subsection{Dataset Statistics and Representations}
We evaluate our framework on two conversation portrait datasets: ViCo and L2L. 
The \textbf{ViCo dataset} comprises 483 face-to-face video clips ranging from 1 to 71 seconds in length, featuring 76 unique listeners and 67 unique speakers. It is strictly partitioned into training ($\mathcal{D}_{train}$), in-domain testing ($\mathcal{D}_{test}$), and out-of-domain testing ($\mathcal{D}_{ood}$) subsets, where identities in $\mathcal{D}_{ood}$ are entirely unseen during training. For ViCo, motion is parameterized using 64-dimensional (64D) expression coefficients coupled with a 6D global head pose (comprising 3D rotation and 3D translation vectors). The acoustic driving signal is extracted using a pre-trained Wav2vec 2.0 encoder, which provides a robust, phonetically-aware acoustic representation. During training, sequence lengths are set to $T=128$.

The \textbf{L2L dataset} is a massive ``in-the-wild'' corpus sourced from YouTube, featuring 72 hours of training data and 95 minutes of testing data across six identities. Unlike ViCo, L2L does not include original video frames; it exclusively provides 3D motion coefficients and pre-extracted audio features. Motion is represented via 50D expression coefficients and a 6D pose (3D jaw rotation and 3D head rotation). Training sequence lengths for L2L are set to $T=64$. 
For both datasets, the autoregressive temporal window size is consistently maintained at $W = 16$ to preserve historical anchoring without inducing excessive computational overhead.

\subsection{Evaluation Metric Definitions}
Because photorealistic 2D avatar rendering is not the primary focus of this work, all quantitative non-verbal motion evaluations are performed directly on the generated 3D mesh representations rather than 2D rendered pixels. We assess our predictions along four distinct axes:
\begin{itemize}[leftmargin=*, nosep]
    \item \textbf{Accuracy:} We compute the point-wise $L_1$ distance between the generated and ground-truth sequences for both expression coefficients and head poses to measure rigid spatial tracking.
    \item \textbf{Realism:} To measure overall distributional similarity, we compute the Fréchet Distance (FD) directly in the expression ($\mathbb{R}^{T \times d_x}$) and pose ($\mathbb{R}^{T \times 6}$) spaces over the full temporal sequence~\cite{ng2022learning}. Furthermore, to capture fine-grained spatial and dynamic realism, we follow \cite{yu2023talking,wang2025diffusion} and report $\text{FID}_{\text{fm}}$ and $\text{FID}_{\Delta \text{fm}}$. Specifically, $\text{FID}_{\text{fm}}$ computes the average Fréchet Distance of the 3DMM coefficients across individual video sequences to evaluate frame-level spatial fidelity. To account for temporal naturalness, $\text{FID}_{\Delta \text{fm}}$ measures the distributional distance of the 3DMM coefficient differences between consecutive frames, thereby explicitly penalizing unnatural inter-frame jitter and dynamic artifacts.
    \item \textbf{Speaker--Listener Correlation:} We report the residual Pearson Correlation Coefficient (rPCC), defined as the $L_1$ discrepancy between the speaker--listener motion correlations computed using generated and ground-truth listener motions~\cite{tran2024dim}. Lower values indicate closer agreement with the reference interaction correlations.
    \item \textbf{Motion Variability and Ground Truth Reference:} To measure the magnitude of kinematic exploration without rewarding erratic noise, we report the average temporal variance (\textbf{var}) across the sequence. Unlike distance metrics where zero is optimal, motion diversity is evaluated by its proximity to the true human distribution ($|\text{var}_{\text{model}} - \text{var}_{\text{GT}}| \downarrow$). The empirical Ground Truth values computed across the test sets are:
    \begin{itemize}[nosep]
        \item \textbf{ViCo $\mathcal{D}_{test}$:} $\text{var}_{\text{exp}} = 0.144$, $\text{var}_{\text{pose}} = 0.047$.
        \item \textbf{ViCo $\mathcal{D}_{ood}$:} $\text{var}_{\text{exp}} = 0.148$, $\text{var}_{\text{pose}} = 0.053$.
        \item \textbf{L2L:} $\text{var}_{\text{exp}} = 0.185$, $\text{var}_{\text{pose}} = 0.013$.
    \end{itemize}
    Values closer to these reference targets indicate natural movement dynamics, penalizing both deterministic under-expression (over-smoothing) and chaotic unconstrained jitter.
\end{itemize}

\subsection{Robotic Embodiment Retraining} 
While the original 64D ViCo coefficients are standard for virtual benchmarking, they lack the explicit anatomical semantics required for physical hardware actuation. Therefore, specifically for the physical embodiment experiments and user study, we retrain REALM alongside all evaluated baselines. To ensure a fair comparison, we pre-process the ViCo training videos using FaceVerse V2~\cite{wang2022faceverse} to extract semantically grounded, 52-dimensional facial blendshape coefficients. These extracted features provide the anatomically consistent basis required for direct and safe inverse kinematic mapping to the Ameca robot's mechanical control values.

\subsection{Two-Stage Training Objectives}
\label{app:training_objectives}

To decouple macro-kinematic stability from high-frequency facial dynamics, REALM is optimized via a two-stage curriculum strategy. In Stage~I, the coarse generator establishes a stable global motion trajectory; in Stage~II, the coarse weights are frozen, and the stochastic refinement module is optimized to recover natural micro-dynamics.

\paragraph{Stage I: Coarse Motion Learning.}
The coarse pathway predicts base motion trajectories $\tilde{\mathbf{m}}_t = [\tilde{\mathbf{x}}_t; \tilde{\mathbf{r}}_t]$ from the reactive context $\mathbf{Z}$. To ensure spatial accuracy while preventing high-frequency jitter and out-of-distribution values, we optimize:
\begin{equation}
\mathcal{L}_{\text{coarse}} = \mathcal{L}_{\text{rec}} + \lambda_v \mathcal{L}_{\text{vel}} + \lambda_b \mathcal{L}_{\text{bnd}},
\label{eq:loss_coarse_full}
\end{equation}
where the loss components are defined as follows:

\begin{enumerate}[leftmargin=*, nosep]
    \item \textbf{Reconstruction Loss ($\mathcal{L}_{\text{rec}}$):} Supervises point-wise spatial tracking across both expression and rigid pose components using an $L_1$ distance:
    \begin{equation}
        \mathcal{L}_{\text{rec}} = \frac{1}{T}\sum_{t=1}^{T} \left( \|\tilde{\mathbf{x}}_t - \mathbf{x}_t\|_1 + \|\tilde{\mathbf{r}}_t - \mathbf{r}_t\|_1 \right),
    \end{equation}
    where $\mathbf{x}_t$ and $\mathbf{r}_t$ denote ground-truth expression and head-pose coefficients, respectively.

    \item \textbf{Velocity Regularization ($\mathcal{L}_{\text{vel}}$):} Enforces smooth inter-frame transitions by penalizing discrepancies in first-order temporal differences:
    \begin{equation}
        \mathcal{L}_{\text{vel}} = \frac{1}{T-1}\sum_{t=2}^{T} \left\| (\tilde{\mathbf{m}}_t - \tilde{\mathbf{m}}_{t-1}) - (\mathbf{m}_t - \mathbf{m}_{t-1}) \right\|_2^2.
    \end{equation}
    The loss weight is set to $\lambda_v = 1.0$, which prevents servo-damaging jerkiness without over-damping conversational responsiveness.

    \item \textbf{Boundary Constraint ($\mathcal{L}_{\text{bnd}}$):} Penalizes predictions that exceed the valid biological motion envelope $[\mathbf{m}_{\min}, \mathbf{m}_{\max}]$, computed from the training distribution:
    \begin{equation}
        \mathcal{L}_{\text{bnd}} = \frac{1}{T}\sum_{t=1}^{T} \left( \left\| \operatorname{ReLU}(\mathbf{m}_{\min} - \tilde{\mathbf{m}}_t) \right\|_2^2 + \left\| \operatorname{ReLU}(\tilde{\mathbf{m}}_t - \mathbf{m}_{\max}) \right\|_2^2 \right).
    \end{equation}
    We set $\lambda_b = 5.0$ as a strict penalty weight to ensure generated trajectories never drift into anatomically impossible or hardware-unsafe configurations.
\end{enumerate}

\paragraph{Stage II: Stochastic Refinement.}
Once Stage~I converges, the coarse generator parameters are frozen. The stochastic refinement module $\mathcal{R}_\theta$ is then trained to generate residual non-rigid expressions $\Delta\mathbf{x}_t = [\mathcal{R}_\theta(\mathbf{Z}^r)]_t$, forming the final prediction $\hat{\mathbf{m}}_t = [\tilde{\mathbf{x}}_t + \Delta\mathbf{x}_t; \, \tilde{\mathbf{r}}_t]$. 

Because standard deterministic reconstruction objectives penalize multimodal variation and induce mean-pose over-smoothing, Stage~II combines a coordinate preservation loss with an adversarial objective:
\begin{equation}
\mathcal{L}_{\text{refine}} = \mathcal{L}_{1} + \lambda_{\text{adv}} \mathcal{L}_{\text{adv}},
\label{eq:loss_refine_full}
\end{equation}
with the balancing weight empirically set to $\lambda_{\text{adv}} = 0.5$.

\begin{enumerate}[leftmargin=*, nosep]
    \item \textbf{Spatial Consistency Loss ($\mathcal{L}_1$):} Preserves coarse trajectory alignment and anchors the refinement within the target expression manifold:
    \begin{equation}
        \mathcal{L}_{1} = \frac{1}{T}\sum_{t=1}^{T} \|\hat{\mathbf{x}}_t - \mathbf{x}_t\|_1 = \frac{1}{T}\sum_{t=1}^{T} \|(\tilde{\mathbf{x}}_t + \Delta\mathbf{x}_t) - \mathbf{x}_t\|_1.
    \end{equation}

    \item \textbf{Adversarial Loss ($\mathcal{L}_{\text{adv}}$):} We employ a temporal convolutional discriminator $D_\phi$ to evaluate both expression sequences $\hat{\mathbf{X}} = \{\hat{\mathbf{x}}_t\}_{t=1}^T$ and their inter-frame velocities $\Delta\hat{\mathbf{X}} = \{\hat{\mathbf{x}}_t - \hat{\mathbf{x}}_{t-1}\}_{t=2}^T$. Using the Least-Squares GAN (LSGAN) formulation for training stability, the discriminator and generator objectives are:
    \begin{align}
        \mathcal{L}_D &= \frac{1}{2} \mathbb{E}_{\mathbf{X} \sim p_{\text{data}}} \left[ (D_\phi(\mathbf{X}) - 1)^2 \right] + \frac{1}{2} \mathbb{E}_{\mathbf{Z}^r} \left[ D_\phi(\hat{\mathbf{X}})^2 \right], \\
        \mathcal{L}_{\text{adv}} &= \mathbb{E}_{\mathbf{Z}^r} \left[ (D_\phi(\hat{\mathbf{X}}) - 1)^2 \right].
    \end{align}
    This adversarial supervision encourages the refinement latent $\mathbf{Z}^r$ to synthesize authentic high-frequency dynamics (e.g., rapid eyelid closures and subtle brow twitches) that match human perceptual characteristics.
\end{enumerate}

\paragraph{Optimization Schedule.}
Both stages use the AdamW optimizer ($\beta_1 = 0.9, \beta_2 = 0.999$, weight decay $10^{-4}$) with a batch size of 512. Stage~I is trained for 120 epochs with an initial learning rate of $10^{-3}$, modulated via a cosine annealing scheduler with a $10\%$ linear warmup. Stage~II is trained for an additional 80 epochs with an initial learning rate of $2 \times 10^{-4}$ for both the refinement generator and the temporal discriminator.

\subsection{Model Capacity and Compute Resources}
Our REALM network was optimized on a single NVIDIA RTX 5090 GPU. Using a batch size of 512, training converges in 200 epochs with an initial learning rate of 1e-3, utilizing the AdamW optimizer and a \texttt{cosine\_schedule\_with\_warmup} scheduler with a warmup ratio of 0.1. The total number of trainable parameters in the REALM framework is strictly 1.50M. Operating at only 0.02G FLOPs per forward pass (using a temporal window size of $W = 16$), our model is substantially more lightweight and computationally efficient than leading state-of-the-art baselines, requiring roughly $4\times$ fewer parameters than ListenFormer (6.13M) and L2L (5.41M).

\subsection{Module Configurations}
All core modules project features into a shared latent dimension of 128 and consistently utilize 4 attention heads across their respective attention mechanisms. Specifically, the Speaker Encoder employs a 2-layer architecture combining Conv1D with a Transformer Encoder. The Speaker-Listener Fusion module operates via a single-layer Transformer Decoder equipped with our Shifted ALiBi. Coarse motion is generated by the Reaction Decoder, which utilizes a heavier 4-layer stack consisting of a Transformer Decoder coupled with an LSTM to ensure stable trajectory regression. Finally, the Refinement Module synthesizes high-frequency micro-dynamics using a single-layer Dilated Conv1D and Transformer Encoder to maintain low-latency inference.

\section{Physical Grounding: Inverse Kinematic Mapping and Control Safety}
\label{app:inverse_mapping}

\paragraph{End-to-End Physical Grounding Pipeline.}
Transforming generated facial coefficients $\hat{\mathbf{m}}_t = [\hat{\mathbf{x}}_t; \hat{\mathbf{r}}_t]$ into physical actuation on the humanoid robot requires addressing morphology differences, mechanical resting offsets, and hardware safety envelopes. We execute this via a four-stage deterministic pipeline:
\begin{equation}
\hat{\mathbf{m}}_t 
\;\xrightarrow{\;\text{Inverse Kinematics}\;}\; \mathbf{q}_t 
\;\xrightarrow{\;\text{Smoothing}\;}\; \tilde{\mathbf{q}}_t 
\;\xrightarrow{\;\text{Relative Calibration}\;}\; \mathbf{q}_t^{\text{calib}} 
\;\xrightarrow{\;\text{Hardware Safety Clamping}\;}\; \mathbf{u}_t.
\label{eq:physical_pipeline_revised}
\end{equation}

\begin{enumerate}[leftmargin=*, nosep]
    \item \textbf{Inverse Kinematic Projection ($\Phi^{-1}$):} Maps the 55D facial representation $\mathbf{c}_t \in \mathbb{R}^{55}$ (comprising 52 semantic blendshapes $\mathbf{x}_t$ and 3D head rotation $\mathbf{r}_t$) to intermediate robot joint space:
    \begin{equation}
        \mathbf{q}_t = \Phi^{-1}(\mathbf{c}_t) = \mathbf{W}\mathbf{c}_t + \mathbf{b}_0,
    \end{equation}
    where $\mathbf{W} \in \mathbb{R}^{d_q \times 55}$ is the morphological coupling matrix encoding synergistic and antagonistic muscle mappings, and $\mathbf{b}_0 \in \mathbb{R}^{d_q}$ denotes the baseline mechanical neutral offset.
    
    \item \textbf{Temporal Smoothing ($\mathcal{S}$):} To eliminate high-frequency coefficient jitter and prevent servo chattering, we apply a Savitzky--Golay filter $\tilde{\mathbf{q}}_t = \mathcal{S}(\mathbf{q}_t)$ across a sliding temporal window of $k=5$ frames.

    \item \textbf{Relative Motion Calibration ($\mathcal{C}$):} To prevent human facial resting posture from biasing the robot's physical configuration, we transfer only relative motion excursions around the robot's pre-calibrated mechanical neutral state $\mathbf{q}_0$:
    \begin{equation}
        \mathbf{q}_t^{\text{calib}} = \mathbf{q}_0 + \Delta \tilde{\mathbf{q}}_t = \mathbf{q}_0 + (\tilde{\mathbf{q}}_t - \tilde{\mathbf{q}}_{\text{ref}}),
    \end{equation}
    where $\tilde{\mathbf{q}}_{\text{ref}}$ is the mean resting posture extracted from the initial neutral frames.

    \item \textbf{Post-Calibration Safety Clamping and Controller Checks:} 
    Because relative offsets can mathematically shift $\mathbf{q}_t^{\text{calib}}$ outside operational limits, the actual executable command $\mathbf{u}_t$ sent to the motor controllers is strictly clamped at the final stage:
    \begin{equation}
        \mathbf{u}_t = \operatorname{clip}\left(\mathbf{q}_t^{\text{calib}}, \, \mathbf{q}_{\min}, \, \mathbf{q}_{\max}\right),
        \label{eq:safety_clamping}
    \end{equation}
    subject to low-level motor rate-of-change constraints $\|\mathbf{u}_t - \mathbf{u}_{t-1}\|_\infty \le \mathbf{v}_{\max}\Delta t$ and current-overload safety cutoffs implemented in the robot firmware.
\end{enumerate}

\subsection{Kinematic Subspace Formulations}
Rather than unconstrained heuristic scaling, the coupling matrix $\mathbf{W}$ decomposes into dedicated anatomical sub-matrices corresponding to independent mechanical subsystems. All angular rotational targets (head orientation and ocular gaze) are strictly parameterized in \textbf{radians}, matching the physical controller specifications.

\paragraph{1. Ocular Dynamics and Gaze Kinematics.}
Eye dynamics differentiate between bilateral blinks, resting aperture, and directional gaze:
\begin{align}
    q_{\text{lid\_upper}}^{\text{side}} &= b_{\text{lid}} - \alpha_{\text{blink}} c_{\text{blink}}^{\text{side}} + \alpha_{\text{wide}} c_{\text{wide}}^{\text{side}}, \quad (\text{side} \in \{\text{L}, \text{R}\}), \\
    q_{\text{gaze}}^{\theta} &= k_{\text{gaze}}^\theta \big[(c_{\text{lookUp}}^{\text{L}} + c_{\text{lookUp}}^{\text{R}}) - (c_{\text{lookDown}}^{\text{L}} + c_{\text{lookDown}}^{\text{R}})\big], \\
    q_{\text{gaze}}^{\phi} &= k_{\text{gaze}}^\phi \big[(c_{\text{lookIn}}^{\text{L}} + c_{\text{lookOut}}^{\text{R}}) - (c_{\text{lookOut}}^{\text{L}} + c_{\text{lookIn}}^{\text{R}})\big],
\end{align}
where $k_{\text{gaze}}^\theta \approx 0.55\text{ rad}$ ($\approx 31.5^\circ$) and $k_{\text{gaze}}^\phi \approx 1.15\text{ rad}$ ($\approx 65.9^\circ$) scale differential eye gaze coefficients directly to the mechanical ocular envelope $[-1.1, 1.1]\text{ rad}$ and $[-2.3, 2.3]\text{ rad}$, respectively.

\paragraph{2. Rigid Head Kinematics.}
Head rotation coefficients are mapped directly in radians from the estimated camera frame to the robot neck coordinate system:
\begin{equation}
    \mathbf{q}_{\text{head}} = \begin{bmatrix} q_{\text{pitch}} \\ q_{\text{yaw}} \\ q_{\text{roll}} \end{bmatrix} 
    = \mathbf{R}_{\text{cam}}^{\text{robot}} \mathbf{c}_{\text{rot}} + \mathbf{b}_{\text{head}},
\end{equation}
where $\mathbf{R}_{\text{cam}}^{\text{robot}} = \operatorname{diag}(1, -1, 1)$ aligns coordinate conventions. The physical hardware limits for head rotation are $[-0.50, 0.50]\text{ rad}$ for yaw ($\approx \pm 28.6^\circ$, informally bounded at $\pm 30^\circ$) and $[-0.50, 0.30]\text{ rad}$ for pitch ($\approx -28.6^\circ \text{ to } +17.2^\circ$).

\paragraph{3. Lower Face: Synergistic and Antagonistic Coupling.}
To prevent opposing servo strain in regions of dense actuation, lip actuators combine synergistic blendshapes additively while penalizing antagonistic muscle activations:
\begin{equation}
    q_{\text{lip}}^{\text{target}} = b_{\text{lip}} + \mathbf{w}_{\text{syn}}^\top \mathbf{c}_{\text{syn}} - \mathbf{w}_{\text{ant}}^\top \mathbf{c}_{\text{ant}},
\end{equation}
where, for example, the lip corner elevator $q_{\text{lip\_corner\_raise}}$ is driven synergistically by smiling ($c_{\text{smile}}$) and inhibited antagonistically by frown ($c_{\text{frown}}$) and lip-press ($c_{\text{press}}$) coefficients. For overlapping bilateral muscle groups such as nasal sneering, non-linear max-pooling avoids compounding strain on localized actuators:
\begin{equation}
    q_{\text{nose\_wrinkle}} = \max(c_{\text{sneer}}^{\text{L}}, \, c_{\text{sneer}}^{\text{R}}).
\end{equation}

\begin{table*}[t]
\centering
\caption{Physical robot actuator channels, control modalities, and absolute hardware limits $[\mathbf{q}_{\min}, \mathbf{q}_{\max}]$ enforced by the final post-calibration safety clamping stage (Eq.~\ref{eq:safety_clamping}). Linear facial actuators operate in normalized control units, while head kinematics and gaze targets operate strictly in radians.}
\vspace{5pt}
\resizebox{\textwidth}{!}{%
\begin{tabular}{lcccccc}
\toprule
\textbf{Actuator Group} & \textbf{Channel Name} & \textbf{Unit} & $\mathbf{q}_{\min}$ & $\mathbf{q}_{\max}$ & \textbf{Neutral } $\mathbf{q}_0$ & \textbf{Primary Driving Blendshapes} \\
\midrule
\multirow{2}{*}{\textbf{Jaw}} 
 & Jaw Pitch & Normalized & $0.00$ & $1.00$ & $1.00$ & Jaw Open ($c_{\text{jawOpen}}$) \\
 & Jaw Yaw & Normalized & $0.00$ & $1.00$ & $0.50$ & Jaw Left/Right ($c_{\text{jawLeft}}, c_{\text{jawRight}}$) \\
\midrule
\multirow{4}{*}{\textbf{Brows \& Eyelids}} 
 & Brow Inner (L/R) & Normalized & $0.00$ & $1.00$ & $0.50$ & Brow Down / Brow Inner Up \\
 & Brow Outer (L/R) & Normalized & $0.00$ & $1.00$ & $0.50$ & Brow Outer Up \\
 & Eyelid Upper (L/R) & Normalized & $-1.00$ & $2.00$ & $1.00$ & Eye Blink / Eye Wide \\
 & Eyelid Lower (L/R) & Normalized & $-1.00$ & $2.00$ & $0.00$ & Eye Squint \\
\midrule
\multirow{2}{*}{\textbf{Ocular Gaze}} 
 & Gaze Pitch ($\theta$) & Radians & $-1.10$ & $1.10$ & $0.00$ & Eye Look Up / Down ($\approx \pm 63.0^\circ$) \\
 & Gaze Yaw ($\phi$) & Radians & $-2.30$ & $2.30$ & $0.00$ & Eye Look In / Out ($\approx \pm 131.8^\circ$) \\
\midrule
\multirow{3}{*}{\textbf{Head Pose}} 
 & Head Pitch & Radians & $-0.50$ & $0.30$ & $0.00$ & Pitch Rotation ($\approx -28.6^\circ \text{ to } +17.2^\circ$) \\
 & Head Yaw & Radians & $-0.50$ & $0.50$ & $0.00$ & Yaw Rotation ($\approx \pm 28.6^\circ$) \\
 & Head Roll & Radians & $-0.30$ & $0.30$ & $0.00$ & Roll Rotation ($\approx \pm 17.2^\circ$) \\
\midrule
\multirow{3}{*}{\textbf{Lower Face}} 
 & Lip Corner Raise (L/R) & Normalized & $0.00$ & $1.00$ & $0.47$ & Smile ($+$), Frown ($-$), Press ($-$) \\
 & Lip Top Raise (L/R/Mid) & Normalized & $0.00$ & $1.00$ & $0.30$ & Upper Lip Up ($+$), Shrug Upper ($-$) \\
 & Nose Wrinkle & Normalized & $0.00$ & $1.00$ & $0.00$ & Bilateral Sneer ($\max$) \\
\bottomrule
\end{tabular}%
}
\label{tab:ctrl_range}
\end{table*}

\section{Structural Analysis of Micro-Dynamics: Blink Extraction}
\label{app:blink_analysis}

A primary challenge in stochastic motion synthesis is ensuring that high-frequency variance translates into anatomically meaningful micro-expressions (e.g., natural eye blinks) rather than unstructured jitter. To achieve this, REALM explicitly avoids injecting random noise directly into the output coordinate space. Instead, the stochastic noise tensor $\boldsymbol{\epsilon} \sim \mathcal{N}(0, \mathbf{I})$ is introduced exclusively within the \textit{latent} feature space ($\mathbf{Z}_c$), where it is rigorously constrained by two dedicated semantic components:
\begin{enumerate}
    \item \textbf{Audio Modulator (FiLM):} The latent noise is conditioned by time-varying scale ($\boldsymbol{\gamma}$) and shift ($\boldsymbol{\beta}$) parameters~\cite{perez2018film}. Because these are derived directly from the speaker's acoustic embeddings, the intensity of the injected variance is explicitly modulated by the acoustic energy, ensuring the stochasticity remains semantically anchored to the conversation's flow.
    \item \textbf{Channel Attention \& Temporal Coherence:} Different facial muscle groups exhibit varying sensitivities to micro-dynamics (e.g., blinks are rapid, whereas jaw movements are smooth). Our Channel Attention mechanism dynamically re-weights specific latent dimensions, while the Self-Attention Encoder enforces global temporal coherence. This pipeline explicitly maps audio-conditioned latent stochasticity into biomechanically structured movements, preventing anatomically decoupled jitter.
\end{enumerate}

\subsection{Methodology and Evaluation Metrics}
To quantitatively validate the structural correctness of the generated micro-dynamics, we conducted an objective analysis of blink events on the ViCo test set. Because standard 3DMM blendshapes frequently entangle eyelid closures with adjacent facial muscle activations (such as cheek raises), relying solely on raw coefficients for blink detection yields noisy, unreliable estimates. To resolve this, we objectively measured the visual output by reconstructing the listener videos using the official PIRenderer and applying the MediaPipe Face Landmarker across the rendered frames. By extracting continuous eye-aspect scores from the generated facial landmarks, we detected precise temporal blink events to evaluate two metrics (summarized in Table~\ref{tab:blink_analysis}):
\begin{itemize}
    \item \textbf{Rate Error ($\Delta$ Rate):} The absolute difference in Blinks per Minute (BPM) between the generated sequence and the Ground Truth (GT).
    \item \textbf{Duration Error ($\Delta$ Duration):} The absolute difference in average blink duration (ms). Unstructured or poorly constrained noise typically fails this metric by producing unnatural flickering or excessively long, zombie-like eye closures.
\end{itemize}

\begin{table}[h]
\centering
\caption{Structural analysis of micro-dynamics on the ViCo dataset. We evaluate the anatomical correctness of generated eye blinks against the Ground Truth (GT) duration of 292.01 ms.}
\vspace{5pt}
\resizebox{0.8\textwidth}{!}{%
\begin{tabular}{lccc}
\toprule
\textbf{Method} & \textbf{$\Delta$ Rate (BPM) $\downarrow$} & \textbf{Duration Delta ($\Delta$ ms) $\downarrow$} & \textbf{Avg Duration (ms)} \\
\midrule
Without refinement & 10.50 & 74.65 & 366.67 \\
Unmodulated noise & 10.39 & 341.32 & 633.33 \\
\textbf{REALM (Ours)} & \textbf{7.40} & \textbf{18.45} & \textbf{273.56} \\
\bottomrule
\end{tabular}%
}
\label{tab:blink_analysis}
\end{table}

\subsection{Analysis of Results}
As demonstrated in Table~\ref{tab:blink_analysis}, omitting the refinement module entirely leads to severe over-smoothing, resulting in the generation of almost zero blinks. Conversely, substituting our module with unmodulated noise acts as an unconstrained stochastic perturbation; while it mathematically inflates motion variance, it fails to synthesize anatomically correct micro-expressions. This configuration produces physically implausible artifacts with severe duration deviations (averaging $>630$ ms, which resembles an unnatural prolonged closure rather than a rapid, lifelike blink). By contrast, REALM successfully recovers these high-frequency dynamics while maintaining strict structural fidelity. It significantly improves the physiological blink rate and yields an anatomical blink duration (273.56 ms) that closely mirrors the natural biomechanics of the Ground Truth (292.01 ms).

\section{Subjective User Study Details}
\label{app:user_study}

While mathematical metrics provide a proxy for motion fidelity, the perceptual quality of conversational dynamics is inherently subjective, particularly when deployed on physical hardware. To rigorously evaluate the real-world viability of our complete pipeline—from digital generation to physical execution via our physical grounding pipeline (Sec.~\ref{sec:robotic-embodiment})—we conducted a subjective Mean Opinion Score (MOS) user study on the physical Ameca humanoid robot.

\subsection{Procedure and Participants} 
We recruited $N=25$ human evaluators with diverse demographic backgrounds. To prevent survey fatigue and ensure high-quality responses, participants were not asked to evaluate the entire dataset. Instead, each evaluator was presented with a randomized subset of 5 distinct conversational sequences sampled from the ViCo test split ($\mathcal{D}_{test}$), resulting in 125 total sequence evaluations. For each sequence (approximately 15 to 30 seconds in length), participants watched side-by-side video comparisons of the physical Ameca robot driven by our method (REALM) and five baseline methods (RLHG, DSPN, L2L, ListenFormer, and UniLS). To strictly prevent evaluation bias and mitigate the "novelty effect" often associated with viewing humanoid robots, the spatial ordering of the methods was entirely randomized and double-blinded for every sequence. Furthermore, to guarantee the rigor of our data, attention-check questions were seamlessly embedded within the study to filter out inattentive participants.

\subsection{Evaluation Metrics} 
Participants were asked to rate the performance of each method on a 5-point Likert scale (1 = Strongly Disagree/Poor, 5 = Strongly Agree/Excellent) across four distinct perceptual axes:
\begin{itemize}
    \item \textbf{Naturalness (Nat.):} Evaluates the overall lifelike quality of the physical robot's movements, specifically penalizing robotic stiffness, unnatural mechanical jitter, or deterministic over-smoothing.
    \item \textbf{Audio-Visual Synchrony (Syn.):} Evaluates how well the robot's reactive listener motions align temporally with the acoustic and semantic cues of the speaker's driving audio.
    \item \textbf{Contextual Appropriateness (App.):} Assesses whether the generated reactions (e.g., smiling, nodding, blinking) are semantically appropriate for the specific context of the conversation, rather than just being random stochastic noise.
    \item \textbf{Overall Preference (Pref.):} A holistic evaluation where participants indicate which robot they would personally prefer to interact with in a real-world, face-to-face scenario.
\end{itemize}

\subsection{Analysis}
As reported in the main text (Table \ref{tab:user_study}), REALM consistently achieved the highest scores across all four metrics. Baseline methods that rely heavily on deterministic generation (such as RLHG) received the lowest scores, as the translated physical motions appeared rigid and mechanically unresponsive. While ListenFormer performed well in contextual appropriateness, evaluators frequently noted a distinct lack of high-frequency micro-dynamics. REALM effectively resolved this; evaluators noted that our audio-conditioned stochastic refinement, successfully preserved by our relative motion calibration step, generated highly natural, synchronized, and engaging physical embodiments without triggering the uncanny valley.

\section{Comparison with Closely Related Methods}
\label{app:comparison_protocol}

\subsection{Modeling choices.}
REALM builds on history-conditioned and generative models of
conversational motion. ARIG~\cite{guo2025arig} combines
autoregressive interaction modeling with diffusion-based motion
prediction. Wang et al.~\cite{wang2025diffusion} investigate
diffusion-based listener generation with hybrid motion modeling
and tailored guidance for head pose and facial expression.
UniLS~\cite{chu2026unils} first learns an autoregressive motion
prior and subsequently introduces dual-track audio conditioning.

REALM investigates the joint use of an explicit temporal
alignment prior, adaptive context weighting, and expression-specific
stochastic refinement. Its delay-centered attention bias favors
speaker representations near a nominal response lag, while
the learned gate weights speaker context and listener history.
The refinement pathway adds audio-conditioned stochastic
expression residuals to the coarse prediction, preserving
the corresponding coarse pose coordinates.
These mechanisms provide task-specific inductive biases for
balancing motion continuity, responsiveness, and local expression
variation.

\subsection{Scope and baseline selection.}
Our quantitative evaluation focuses on listener expression
and pose coefficients generated from speaker audio and
listener motion history. Comparisons therefore require
compatible motion representations, conditioning inputs,
and evaluation procedures.

ARIG~\cite{guo2025arig} uses a LivePortrait motion
representation and incorporates audio and visual motion
signals from both conversation participants. Its
coefficient-based motion evaluation involves reconstructing
3DMM parameters from rendered videos.
Wang et al.~\cite{wang2025diffusion} combine explicit motion
generation with implicit motion refinement and video
synthesis, conditioned on speaker audio and head motion.
A controlled comparison with these methods would require
aligning their conditioning inputs and motion-evaluation
pipelines with our coefficient-based protocol.

As of 26 September 2026, we could not locate publicly
available training and inference implementations for these
two methods through their official project resources.
Consequently, we do not report reproduced results for
ARIG or Wang et al. under our evaluation protocol.
We discuss their modeling choices qualitatively and
recognize their evaluation under a compatible protocol
as an outstanding comparison.

UniLS~\cite{chu2026unils} is included in our quantitative
evaluation. Our empirical conclusions are restricted to
the methods and configurations evaluated in the reported
experiments.

\end{document}